\documentclass[letterpaper, 10 pt, conference]{IEEEtemplate/ieeeconf}

\IEEEoverridecommandlockouts                              
\usepackage{amsmath,amssymb,amsfonts}
\usepackage{url}
\usepackage{algorithm}
\usepackage{cite}
\usepackage{graphicx}
\usepackage{textcomp}
\usepackage{xcolor}
\usepackage{booktabs}
\usepackage{makecell}
\usepackage{algpseudocode}
\usepackage{multirow}
\usepackage{dsfont}

\newcommand{\calB}{{\cal B}}
\newcommand{\calC}{{\cal C}}
\newcommand{\calD}{{\cal D}}
\newcommand{\calE}{{\cal E}}
\newcommand{\calF}{{\cal F}}

\newcommand{\calK}{{\cal K}}

\newcommand{\calN}{{\cal N}}

\newcommand{\calP}{{\cal P}}

\newcommand{\calU}{{\cal U}}

\newcommand{\calX}{{\cal X}}

\newcommand{\bfu}{\mathbf{u}}

\newcommand{\bfx}{\mathbf{x}}
\newcommand{\bfy}{\mathbf{y}}

\newcommand{\bfR}{\mathbf{R}}

\newcommand{\bfX}{\mathbf{X}}
\newcommand{\bfY}{\mathbf{Y}}

\newcommand{\bbC}{\mathbb{C}}

\newcommand{\bbE}{\mathbb{E}}

\newcommand{\bbP}{\mathbb{P}}

\newcommand{\bbR}{\mathbb{R}}

\DeclareMathOperator{\Var}{Var}
\DeclareMathOperator{\erf}{erf}

\newcommand{\VD}[1]{}
\newcommand{\VDm}[1]{}

\newcommand{\MASm}[1]{}
\newcommand{\DADEE}{DADEE}

\newcommand{\teq}{\triangleq}

\newtheorem{theorem}{Theorem}
\newcommand{\CBC}{\mbox{CBC}}

\title{\LARGE \bf
\DADEE: Well-calibrated uncertainty quantification in neural networks for barriers-based robot safety
}

\author{Masoud Ataei$^{1}$ and Vikas Dhiman$^{2}$
\thanks{*This material is based upon work supported by the National Science Foundation under Grant No 2218063.}
\thanks{$^{1}$Masoud Ataei and $^{2}$Vikas Dhiman are with Electrical and Computer Engineering,
        University of Maine, Orono, ME, USA
      {\tt\small \{masoud.ataei,vikas.dhiman\}@maine.edu}}%
}

\begin{document}

\maketitle
\thispagestyle{empty}
\pagestyle{empty}
\begin{abstract}

Uncertainty-aware controllers that guarantee safety are critical for safety critical applications. Among such controllers, Control Barrier Functions (CBFs) based approaches are popular because they are fast, yet safe. However, most such works depend on Gaussian Processes (GPs) or MC-Dropout for learning and uncertainty estimation, and both approaches come with drawbacks: GPs are non-parametric methods that are slow, while MC-Dropout does not capture aleatoric uncertainty. On the other hand, modern Bayesian learning algorithms have shown promise in uncertainty quantification. The application of modern Bayesian learning methods to CBF-based controllers has not yet been studied. We aim to fill this gap by surveying uncertainty quantification algorithms and evaluating them on CBF-based safe controllers.

We find that model variance-based algorithms (for example, Deep ensembles~\cite{lakshminarayanan2017simple}, MC-dropout~\cite{gal2016dropout}, etc.) and direct estimation-based algorithms (such as DEUP~\cite{lahlou2021deup}) have complementary strengths.
Algorithms in the former category can only estimate uncertainty accurately out-of-domain, while those in the latter category can only do so in-domain. 
We combine the two approaches to obtain more accurate uncertainty estimates both in- and out-of-domain. As measured by the failure rate of a simulated robot, this results in a safer CBF-based robot controller.
\end{abstract}
\section{Introduction}

Using machine learning in robotic systems introduces uncertainty. This uncertainty needs to be quantified accurately for safety-critical robotic applications. 
We build upon recent works~\cite{dhiman2023controlbarriers,long2022safe,castaneda2022probabilistic, cheng2019end, fan2020bayesian} which provide a CBF-based control synthesis approach, an approach that guarantees safety, provided that the learned system dynamics' uncertainty is accurately estimated. 
However, the use of modern Bayesian learning methods (for example, SWAG~\cite{maddox2019simple}, Laplace approximation~\cite{daxberger2021laplace}, etc.) applied to CBF for safe robot navigation is underexplored. Existing works~\cite{dhiman2023controlbarriers,long2022safe,castaneda2022probabilistic} either use Gaussian processes (GP)~\cite{williams2006gaussian} or MC-Dropout~\cite{gal2016dropout} to learn the system dynamics with uncertainty estimation. Both of these approaches come with drawbacks: GP is known to be slow at inference time, which grows proportionally to the cube $O(n^3)$ of the training dataset size, $n$. 
On the other hand, we find that the MC-Dropout fails to learn the \emph{aleatoric} uncertainty (Sec~\ref{sec:1-D-regression-experiment}).
Aleatoric uncertainty, unlike \emph{epistemic} uncertainty, is irreducible uncertainty due to sensor noise or inherent uncertainty in the process generating the data. Epistemic uncertainty, on the other hand, is uncertainty due to the limitation of the modeler's knowledge, for example due to choosing the incorrect model class or due to the insufficient amount of data collected. 


We build on these works to find a balance between efficiency and accuracy for uncertainty estimation.
We briefly survey the literature for uncertainty estimation and chose the following algorithms to represent the various categories of different Bayesian learning approaches: Deep Ensembles~\cite{lakshminarayanan2017simple}, Anchored Ensembles~\cite{pearce2020uncertainty}, Stochastic Weighted Averaging-Gaussian (SWAG)~\cite{maddox2019simple}, Laplace approximation~\cite{daxberger2021laplace}, Maximum Likelihood with Learned Variance (MLLV)~\cite{lakshminarayanan2017simple}, and Direct Epistemic Uncertainty Prediction (DEUP)~\cite{lahlou2021deup}. We begin by comparatively evaluating these algorithms in a 1-D regression experiment. This allows us to test the algorithms in isolation and gives us a high degree of control over both in-domain and out-of-domain sampling. Our experiments find that, of these algorithms, Anchored Ensembles most accurately estimates out-of-domain (OOD) uncertainty with low resource usage, while direct estimation-like approaches~\cite{lahlou2021deup, lakshminarayanan2017simple} are the best at calculating in-domain uncertainty. We combine both approaches to create a new method that we call the Direct Aleatoric and Deep Ensemble-based Epistemic~(\DADEE{}) uncertainty estimator. We apply this method to an uncertainty-aware CBF-based safe controller for robot navigation.
\begin{figure*}%
\centering
\includegraphics[width=\textwidth,trim=0.3in 0.45in 0.1in 0.38in,clip]{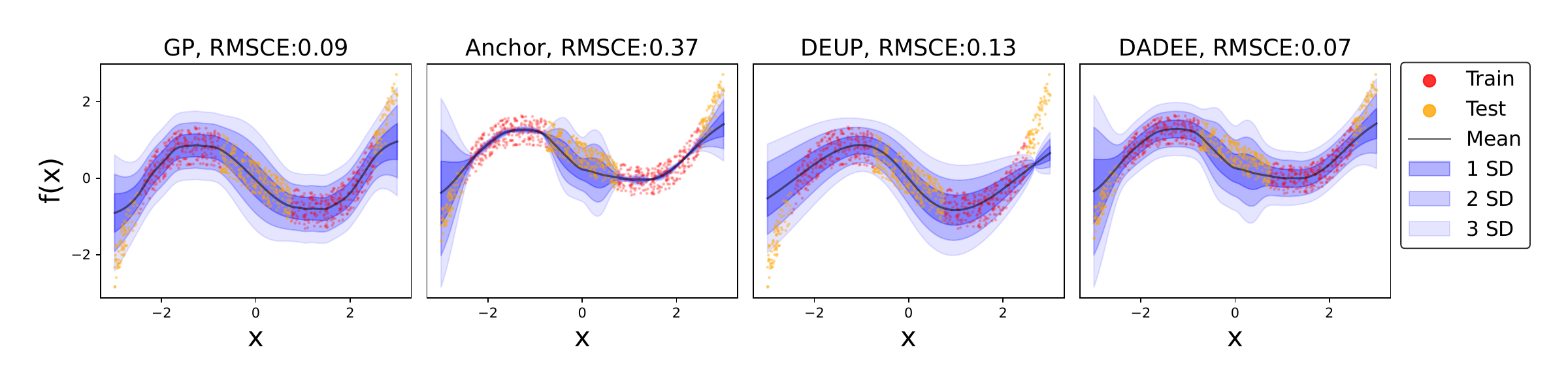}%
    \caption{The posterior distribution prediction for GP, Anchored Ensemble, DEUP, and DADEE. Anchored ensemble estimates uncertainty accurately out-of-domain, increasing the number of models results in more accurate estimation, while DEUP does so in-domain. GP and DADEE recover the true distribution correctly in both out-of-domain and in-domain. }
    \label{fig:1-D Example results}
\end{figure*}%
We then reevaluate the aforementioned algorithms on the intended application of safe robot navigation. We do so by applying them to an uncertainty-aware CBF-based safe controller for robot navigation. We use a robotic simulation setup similar to that provided by Dhiman et al.~\cite{dhiman2023controlbarriers} which measures the safety-risk of the robot. In this simulation setup, a robot starts with unknown dynamics and a large prior uncertainty. As the robot moves around collecting data and learning from it, the learned system dynamics approaches the true dynamics, and uncertainty correspondingly decreases. It is important to estimate uncertainty accurately: if uncertainty is underestimated, the robot risks crashing into obstacles; if overestimated, the robot controller will be excessively risk-averse, limiting its capabilities; for example, refusing to venture into spots with multiple close obstacles. Clearly, overestimating uncertainty is preferable to underestimating, but accurate estimation is ideal.

We make the following \textbf{contributions} in this paper. (1) We survey various uncertainty quantification algorithms and evaluate them under both in-domain and out-of-domain conditions. We find that model variance-based algorithms and direct estimation-based algorithms have complementary strengths: the former estimates uncertainty accurately out-of-domain, while the latter does so in-domain (Sec~\ref{sec:1-D-regression-experiment}). (2) We combine the two to obtain an accurate estimate both in- and out-of-domain. As compared to the surveyed literature, our combined algorithm provides the most accurate uncertainty estimates. (3) We evaluate the uncertainty algorithms including our combined uncertainty estimation in an uncertainty-aware CBF-based safe robot controller and measure the system for safety risk in simulation. To the best of our knowledge, this is the first paper to do so. (Sec~\ref{sec:robot-simulation}). As compared to the surveyed literature, we find our uncertainty-aware controller (\DADEE{}) to be the safest. For the reproducibility of our experiments, we open source\footnote{Code available at \url{https://github.com/Ataei67/DADEE}} our code and models.

\section{Background}


\subsection{Uncertainty estimation in deep learning}

Uncertainty estimation in deep learning has been interpreted under both frequentist and Bayesian learning frameworks~\cite{abdar2021review}. In this paper, we adopt the Bayesian interpretation because it provides a unifying framework for estimating probability distribution over its outputs, as opposed to just a point estimate. Moreover, many works initially proposed as frequentist have been later reinterpreted as Bayesian, including MC-Dropout~\cite{gal2016dropout} and Deep ensembles~\cite{hoffmann2021deep}, both of which we evaluate in this paper. 

The broad framework of parameteric Bayesian learning (BL) is summarized as follows.
In BL, given a training dataset, $\calD^n = \{(\bfx_1, \bfy_1), \dots (\bfx_n, \bfy_n)\}$ assumed to be sampled i.i.d. from an unknown joint distribution $\bbP(\bfX, \bfY)$, one seeks to estimate the distribution $\bbP(\bfy^*|\bfx^*, \calD^n)$ at a given test point $\bfx^*$.  BL limits the distribution search to a hypothesis class of models $\bbP(\bfy_i|\bfx_i, \theta) \triangleq f_\theta(\bfy_i|\bfx_i) \in \calF$ that are parameterized by a finite-dimensional parameter vector $\theta \in \Theta \subseteq \bbR^d$\footnote{Non-parameteric methods like Gaussian Processes can handle infinite dimensional feature spaces by using finite dimensional combinations of kernel function ~\cite[p132]{williams2006gaussian}. }. The first step is to estimate a distribution over the parameters $\theta$ using the training data $\calD^n$,
\begin{align}
    \bbP(\theta|\bfY, \bfX) &= \frac{\bbP(\bfY|\bfX, \theta) \bbP(\theta)}{\bbP(\bfY|\bfX)}
    &=\prod_{i \in \calD^n}  \frac{\bbP(\bfy_i|\bfx_i, \theta)\bbP(\theta)}{Z},\notag
\end{align}%
where $Z$ is the normalization factor. Often, $\bbP(\theta|\calD)$ cannot be computed analytically~\cite{graves2011practical} and is expensive to sample. Instead, BL approximates $\bbP(\theta|\calD)$ using another parametric distribution, $Q_\beta(\theta)\approx \bbP(\theta|\calD)$.
BL then proceeds by estimating a distribution over $\bfy^*$ for a given test data point $\bfx^*$ by marginalizing over all possible $\theta \in \Theta$,
\begin{align}
    \bbP(\bfy^*|\bfx^*,\calD) &= \frac{1}{Z'} \int_{\theta \in \Theta} \bbP(\bfy^*|\bfx^*, \theta) Q_\beta(\theta) d\theta.
    \label{eq:bayes-marginal}
\end{align}%
In large and complex parametric models, it is intractable to compute the marginal $\int_{\theta \in \Theta}$ over all possible parameters. For that reason, approximation-based methods are often used instead.
\subsubsection{Approximate Bayesian learning}
We discuss broad categories of such approximations:
\paragraph{Variational inference (VI)}  VI finds the parameters $\beta$ of the approximate distribution $Q_\beta(\theta)$ by minimizing the variational free energy which is the KL-divergence between the posterior probability $\bbP(\theta|\calD)$ and its approximation $Q_\beta(\theta)$~\cite{graves2011practical},
\begin{align}
\beta^* &= \arg~\min_{\beta} D_{KL}(Q_\beta(\theta) \| \bbP(\theta|\calD)) 
\label{eq:variational} \\
&= \arg~\min_{\beta}\bbE_{\theta\sim Q}[-\ln\bbP(\bfY|\bfX, \theta)] + D_{KL}(Q_\beta(\theta)\| \bbP(\theta)). \notag
\end{align}
The first term is interpreted as \emph{data loss} or \emph{negative log likelihood} and the second term regularizes the approximate distribution to be close to the prior.
In practical implementation of VI~\cite{graves2011practical}, $Q_\beta(\theta)$ is assumed to be a tractable distribution like diagonal Gaussian distribution~\cite{graves2011practical},  or an inverse-Gamma distribution~\cite{amini2020deep}.

\paragraph{Direct estimation of uncertainty}
\label{sec:direct-estimators}
In direct estimators, the uncertainty is estimated from the data loss term or negative log likelihood term of \eqref{eq:variational}, rather than the regularizer term. We consider two kinds of the methods in this category, Maximum Likelihood with Learned Variance (MLLV)~\cite{lakshminarayanan2017simple,nix1994meanAndVariance} and Direct Epistemic Uncertainty Prediction (DEUP)~\cite{lahlou2021deup}.

In MLLV, the likelihood is assumed to be a Gaussian whose mean $\mu_\theta(\bfx)$ and variance $\sigma^2_\theta(\bfx)$ are neural networks that are learned jointly using the maximization of a single likelihood function,
\begin{align}
    -\ln\bbP(\bfy_i|\bfx_i, \theta) =\frac{\ln \sigma^2_\theta(\bfx_i)}{2} + \frac{(\bfy_i -\mu_\theta(\bfx_i))^2}{2\sigma^2_\theta(\bfx)} .
\end{align}

DEUP also estimates the mean and variance networks, but the two networks are trained in stages. In the first stage, only the mean network is trained. In the second stage, the variance network is trained on a new dataset collected from the errors made by the mean network. Note that our DEUP approach differs in many details from Lahlou et al.~\cite{lahlou2021deup}, for example, we are not using DEUP for estimating epistemic uncertainty (by subtracting aleatoric uncertainty). However, the basic idea of two-staged direct estimation remains the same.

\paragraph{Monte-Carlo approaches}
Monte-Carlo approaches can be interpreted as Bayesian inference where the parametric distribution $Q_\beta(\theta)$ is approximated by sum of direct delta distributions~\cite{hoffmann2021deep},
 $Q_\beta(\theta) = \frac{1}{L}  \sum_{l=L} \delta(\theta - \beta_l)$, over points $\{\beta_l\}$ called the ensembles. The predictive distribution over a test point $\bfx^*$, is obtained by Monte Carlo integration over the delta distributions,
     $\bbP(\bfy^*|\bfx^*, \calD) = \frac{1}{L}\sum_{l=1}^L f_{\beta_l}(\bfy|\bfx^*)$.
     
We discuss a few examples of the Monte-Carlo approaches for Bayesian learning. Deep ensembles~\cite{lakshminarayanan2017simple}, Anchored ensembles~\cite{pearce2020uncertainty}, MC-Dropout~\cite{gal2016dropout} and SWAG~\cite{maddox2019simple}.

i) Deep Ensembles~\cite{lakshminarayanan2017simple} initialize multiple neural networks with random weights and train them on random subparts of the training data in random order. Due to this randomness, Deep Ensembles obtain different local minima over the weights even when they are trained on the same loss on the training data. The variance of the different network weights is used as an uncertainty estimate. 

ii) Anchored ensembles~\cite{pearce2020uncertainty} model the prior distribution $\bbP(\theta)$ by a Gaussian distribution that is then approximated by anchor weights $\theta_a$ sampled from $\theta_a\sim\bbP(\theta)$. This anchor weight is then used in an L2 regularizer,
\begin{equation} \label{eq:MSELoss} 
  \beta^*_l =\arg~\min_{\beta_l} \frac{1}{|\calD|}\sum_{i \in \calD} (\bfy_i-f_{\theta_a}(\bfx_i))^2+\frac{\lambda}{|\calD|} (\beta_l-\theta_a)^2,
\end{equation}
where $\lambda$  is the regularization factor.

iii) Gal and Gahremani~\cite{gal2016dropout} showed that MC-Dropout~\cite{srivastava2014dropout,gal2016dropout} in neural networks can be used for uncertainty estimation and demonstrated its equivalence with Gaussian Processes. In MC-Dropout, a certain percentage of activations are made zero so that the network is trained to make correct predictions using only the rest of the subnetwork. Then, at inference time, by randomly dropping different sets of activations, one obtains ensembles of network that can be used for approximating posterior distribution. MC-Dropout is similar to Deep Ensembles in that a new network in the ensemble is produced every time a particular subset of activations are made zero.

iv) In SWAG~\cite{maddox2019simple}  the ensembles are selected after the neural network training has converged. Both SWAG and Stochastic Weight Averaging (SWA)~\cite{izmailov2019averaging} run additional training iterations with a large learning rate to let the stochastic gradient descent trajectory ``bounce around'' the local minimum where the network converged. A fixed number of weights are saved from this trajectory, which is then used to fit a Gaussian distribution over the weights of neural networks taken as an uncertainty estimate. 




\paragraph{Point estimators} Point estimators do not estimate uncertainty, but they are the most commonly used modes of neural networks. They can be interpreted as Bayesian inference, where the approximate distribution $Q_\beta(\theta)$ is a direct delta distribution  $Q_\beta(\theta) = \delta(\theta = \theta^*)$ at the point estimate. There are two variations of point estimation, maximum-a-posteriori (MAP) and maximum likelihood estimate (MLE). In MAP, the mode of the posterior distribution is estimated, 
\begin{align}
    \theta^*_{\text{MAP}} &=  \arg~\max_{\theta \in \Theta} \prod_{(\bfx_i, \bfy_i) \in \calD^n}  \bbP(\bfy_i|\bfx_i, \theta) \bbP(\theta).
\end{align}%
Whereas in MLE, it is further assumed that there is a uniform prior over the parameters and input data samples,
\begin{align}
    \theta^*_{\text{MLE}}
    &=\arg~\max_{\theta \in \Theta} \prod_{(\bfx_i, \bfy_i) \in \calD^n}  \bbP(\bfy_i|\bfx_i, \theta) .
\end{align}%
MLE is often favored due to its simplicity and low computational requirements, but its drawback is that it relies heavily on the assumption of a uniform prior. MAP is preferable when specific prior knowledge about the parameters is available, as it can use this knowledge to provide more accurate results.

\paragraph{Laplace Approximation} LA approximates the posterior distribution $Q_\beta(\theta)$ as a local Gaussian near the $\theta^*_{\text{MAP}}$ using the Laplace as a positive definite approximation of the Hessian near the minimum of the negative log posterior~\cite{daxberger2021laplace},
\begin{align}
    Q_\beta(\theta) = \calN(\theta; \theta^*_{MAP}, \Sigma^*),\quad \Sigma \triangleq (-\nabla_\theta^2 \ln \bbP(\bfy_i | \bfx_i, \theta) )^{-1} \notag. 
\end{align}%
The predictive distribution over outputs is given by,
\begin{align}
\bbP(\bfy^*|\bfx^*, \theta) = \calN\left(\bfy;
f_{\theta_{MAP}}(\bfy|\bfx), 
J_f(\bfx^*)^\top \Sigma J_f(\bfx^*)\right).
\end{align}%
This is not an exhaustive survey of all uncertainty estimation algorithms, but our aim is to cover a representative sample of broad categories of such techniques. For exhaustive surveys of Bayesian learning and uncertainty estimation, please refer to~\cite{gawlikowski2023survey, abdar2021review, psaros2023uncertainty}.



\begin{table*}
    \centering
    \caption{Bayesian Learning evaluated on a 1-D regression problem (Sec.~\ref{sec:1-D-regression-experiment}) DEUP~\cite{lahlou2021deup} has the lowest in-domain RMSCE, and Anchored Ensembles the lowest OOD RMSCE. Our method DADEE has the lowest overall RMSCE. MSLL score of DADEE also is lowest among the all deep models. All models trained for 1000 epochs with learning rate of 0.0001 and batch-size equal to 20.}
    \label{tab: Results of Bayesian Learning Algorithms}
\scalebox{0.9}{
\begin{tabular}{lrrrrrrrrrrr}        
        \toprule
        Model & \makecell{Train \\ time(s)} & \makecell{Infr. \\ time(s)} & MSE & MSLL & \makecell{In-domain \\ MSLL} & \makecell{OOD \\ MSLL} & RMSCE & \makecell{In-domain \\ RMSCE} & \makecell{OOD \\ RMSCE} &	Mem.(bytes) & \makecell{ Number \\ of params }\\       
        \midrule
        GP~\cite{williams2006gaussian}        & 0.470 & 0.042 & 0.093 & \textbf{0.465} & \textbf{0.173} & \textbf{0.895} & 0.097 & 0.069 & 0.152 &  1052676 & 2 \\
        GP\_Sparse~\cite{gpy2014} & 2.717 & \textbf{0.015} & 0.093 & \textbf{0.465} & \textbf{0.173} & \textbf{0.895} & 0.097 & 0.069 & 0.152 & 10816 & 2 \\
        \midrule[0.05pt]
        SWAG~\cite{maddox2019simple}       & \textbf{0.021} & 0.051 & 0.144 & 8.222&	2.288&	20.129& 0.440& 0.433& 0.511 & 18772 & 361 \\
        MC-Dropout~\cite{gal2016dropout}     & 0.042 & 0.021 & 0.137 & 4.606&	0.818&	10.017&        	0.373&	0.285&	0.510 & 1732 & 433 \\
        LA~\cite{daxberger2021laplace}         & 0.041 & 0.238 & 0.107 & 1.948&	1.490&	2.623&         	\textbf{0.336} & \textbf{0.386} & 0.263 & 522728 & 361 \\
        Ensemble~\cite{lakshminarayanan2017simple}        & 0.189 & 0.008 & \textbf{0.097} & 6.594&	2.785&	12.212&  0.516 & 0.544 & 0.474 & 7220 & 1805 \\
        Anchored Ensembles~\cite{pearce2020uncertainty}       & 0.191 & \textbf{0.004} & \textbf{0.097} & 1.891&	2.228&	 \textbf{1.393} & 0.371 & 0.491 & \textbf{0.196} & 14440 & 1805 \\
        \midrule[0.05pt]        
        MLLV~\cite{lakshminarayanan2017simple}   & \textbf{0.038} & \textbf{0.002} & 0.263 & \textbf{0.845} & \textbf{0.205} & \textbf{1.789} & 0.169 & 0.114 & 0.261 & 2888 & 722 \\
        DEUP~\cite{lahlou2021deup}    & \textbf{0.038} & \textbf{0.002} & \textbf{0.107} & 2.376&	0.305&	5.431 & \textbf{0.133} & \textbf{0.067} & \textbf{0.238} & 2888 & 722 \\
        \midrule[0.05pt]        
        MLLV + Anchored  & 0.263 & 0.005 & 0.239 & 0.695&	0.195&	1.432&  0.150 & 0.122 & 0.199 & 15884 &  2166 \\
        DADEE (DEUP + Anchored)  & \textbf{0.176} & \textbf{0.005} & \textbf{0.097} & \textbf{0.473} & \textbf{0.185} & \textbf{0.898} & \textbf{0.069} & \textbf{0.066} & \textbf{0.080} & 15884 & 2166 \\        
        \bottomrule
    \end{tabular}
    }
\end{table*}

\subsection{Control Barrier Funcitons (CBFs)}
We denote the state of a robot as $\bfx \in \calX \subseteq \bbR^n$, and the control input as $\bfu \in \calU \subseteq\bbR^m$. We assume the system evolves with unknown control affine system dynamics,
\newcommand{\ubfu}{\underline{\bfu}}%
\begin{align}
    \dot{\bfx} = f(\bfx) + G(\bfx)\bfu = F(\bfx)\begin{bmatrix}1 &\bfu \end{bmatrix}^\top = F(\bfx) \ubfu.
    \label{eq:system-dynamics}
\end{align}%
A safe control synthesis is the problem of finding a sequence of control inputs $\bfu_{1:k} = \{\bfu_1, \dots, \bfu_k\}$ that  minimizes the cost of taking control inputs $J(.,.)$ while keeping the states $\bfx(t)$ in a given safe set  $\calC_{\text{safe}}$ at all times:
\begin{align}
\label{eq:General_opt}
  \min_{\bfu_{1:k}}\,\,J(\{\bfx(t)\}_{t=0}^{T}, \bfu_{1:k})\,\,
  \text{s.t. }\,\,\bfx(t) \in \calC_{\text{safe}}, \forall t \in [0, T].
\end{align}%

There are multiple control strategies to guarantee the safety of a robot such as Hamilton-Jacobi Reachability (HJR)~\cite{mitchell2005time}, Model Predictive Control (MPC)~\cite{garcia1989model}, Control Barrier Function (CBF)~\cite{mehra2015adaptive}, etc. 
In this paper, we focus on CBF-based controllers because CBF is much faster than HJR or MPC. CBF allows us to focus on a single time step, which reduces computational costs. However, HJR and MPC can be combined with CBF when more computational resources are available~\cite{zeng2021safety, tonkens2022refining, choi2021robust}.

CBF-based approaches ensure the safety of a dynamical system by requiring a continuous function, called the control barrier function $h \in \bbC^1(\calX, \bbR)$, whose zero superlevel set is the safety set, 
$\calC_{\text{safe}} = \{\bfx \in \calX | h(\bfx) \ge 0 \}$. Control barrier function and their application to system safety have been studied extensively, since their introduction by Ames et al.~\cite{ames2016control}. Recently, they have been extended to work on higher-relative degree systems and with uncertain system dynamics~\cite{dhiman2023controlbarriers,long2022safe,castaneda2022probabilistic}. We review some important results.
\begin{theorem}[Control barrier condition~\cite{ames2019CBF,dhiman2023controlbarriers}]
 Consider a control barrier function $h \in \bbC^1(\calX, \bbR)$ whose zero superlevel set is the safety set, 
$\calC_{\text{safe}} = \{\bfx \in \calX | h(\bfx) \ge 0 \}$. Additionally, assume that $\nabla h(\bfx) \ne 0$ for all $\bfx$ when $h(\bfx) = 0$. Then any Lipchitz continuous policy $\pi(\bfx) \in \{\bfu \in \calU \mid \CBC(\bfx, \bfu) \ge 0\}$ render the system \eqref{eq:system-dynamics} safe with respect to $\calC_{\text{safe}}$, where 
\begin{equation}
    \CBC(\bfx, \bfu) \triangleq \nabla_\bfx^\top h(\bfx)F(\bfx) \ubfu + \alpha(h(\bfx)),
    \label{eq:CBCdef}
\end{equation}
and $\alpha$ is an extended class $\calK_\infty$ function. We call the constraint $\CBC$ as the control barrier condition.
\end{theorem}

\begin{theorem}[Prop~4,5\cite{dhiman2023controlbarriers}]
    For the system dynamics in~\eqref{eq:system-dynamics} with the following additional assumption: (a) $F(\bfx)$ is a Gaussian stoachastic process, (b) $F(\bfx)$  is Lipschitz continuous with probabilty $1-\delta_L$; and for a control barrier function, $h(\bfx) \in \bbC^1(\calX, \bbR)$ with relative degree 1, if the Lipchitz continuous control policy $\pi(\bfx_k)$ is determined at time $t_k$, from the following SOCP (Second order cone programming) optimization problem, then the system trajectory \emph{stays in the safe set} $\calC_{\text{safe}}= \{\bfx \in \calX | h(\bfx) \ge 0 \}$ with probability $p_k = \tilde{p}_k(1-\delta_L)$; with a margin $\zeta \ge 0$ and for time duration $\tau_k \le O(\log(\frac{\zeta}{\|\dot{\bfx}_k\|}))$, if $\pi(\bfx_k)$ is given by,
    {\small
    \begin{align}
      \arg\,\inf_{\bfu_k \in \calU}  \|\bfR( \bfx_k) {\bfu_k}\|_2^2,\,
      \text{s.t. }\bbP(\mbox{CBC}_k \ge \zeta|\bfx_k, \bfu_k) \ge \tilde{p}_k
        \label{eq:CBF_opt},
    \end{align}%
    }%
    where $\|\bfR(\bfx_k)\bfu_k\|_2^2$ is the quadratic control cost.
    The probabilistic safety constraint $\bbP(\mbox{CBC}_k \ge \zeta|\bfx_k, \bfu_k) \ge \tilde{p}_k$ is satisfied when the following SOCP constraint is satisfied,
    \begin{align}
            \bbE[\mbox{CBC}_k]& - \zeta - c_p(\tilde{p}_k) \sqrt{\Var[\mbox{CBC}_k]} \ge 0,
            \label{eq:socp-condition}
    \end{align}
    where $c_p(\tilde{p}_k) = \sqrt{2}\erf^{-1}(2\tilde{p}_k - 1)$; with $\erf$ being the Gauss error function.
    \label{thm:socp-controller}
\end{theorem}
The SOCP condition~\eqref{eq:socp-condition} is interpreted as maintaining the $\mbox{CBC}$ to be away from the zero with a margin that is proportional to the standard deviation. The time duration $\tau_k$ for which the system is guaranteed to be safe depends on Lipchitz continuity of the system dynamics, Lipchitz continuity of $h(\bfx)$, the margin $\zeta$ and $\|\dot{\bfx}_k\|$~\cite[Prop.~5]{dhiman2023controlbarriers}.

The \emph{pointwise feasibility} of CBF-SOCP \eqref{eq:socp-condition} is analyzed in~\cite{castaneda2021pointwise}. Equation \eqref{eq:socp-condition} can become infeasible when uncertainty $\Var[\CBC_k]$ grows faster than $\bbE[\CBC_k]$ for all $\bfu_k$.  
In cases when the problem becomes infeasible, we revert to considering safety just in expectation. Designing learning algorithms that keep uncertainty bounded is left for future work.

Note that Thm.~\ref{thm:socp-controller} only guarantees safety for a single time step. 
For guaranteeing long term safety, the above CBF-SOCP controller has to be paired with a long horizon planner or with HJR and MPC~\cite{zeng2021safety, tonkens2022refining, choi2021robust}.


In literature, stochastic control barrier functions have been well explored~\cite{taylor2020learning, buch2021robust, li2023robust, long2022safe}. Various sources of uncertainties have been considered in the literature. 
Tayler et al.~\cite{taylor2020learning} learn the uncertainty of the time derivative of the barrier function. They used this uncertainty to update the safety constraint directly. This approach must learn the uncertainty of different CBFs separately and can only be used for first-relative degree systems. Uncertainty in control inputs studied in~\cite{buch2021robust}. Li et al.~\cite{li2023robust} consider a linear noise and learn the CBF and the uncertain CBF using GP. And, ~\cite{long2022safe} shows that any probabilistic CLF-CBF can be formulated as SOCP.


\section{Methodology}
\paragraph*{\DADEE{}: Direct Aleatoric with Deep Epistemic Ensembles}
We begin by comparatively evaluating uncertainty estimation algorithms in a
1-D regression experiment. The details of these comparisons are in Sec~\ref{sec:1-D-regression-experiment}.
Our 1-D experiments found that in-domain (aleatoric) uncertainty was most accurately estimated by the DEUP algorithm, while out-of-domain (epistemic+aleatoric) uncertainty was most accurately estimated by Anchored Ensembles. We combine these two approaches into a new algorithm, which we call Direct Aleatoric with Deep Epistemic Ensembles (\DADEE) whose pseudocode is provided in Algorithm~\ref{alg:Train_DEE}.   
In our combined approach, we train $L$ networks, $\beta_l, l=1,\dots, L$, using the Anchored Ensembles~\cite{pearce2020uncertainty}. During training, we collect the prediction errors of the network estimates' mean. Then, another network $\beta_v$ is trained on these collected error estimates similar to DEUP described in Sec.~\ref{sec:direct-estimators}. The latter network is the direct estimator of in-domain uncertainty. The pseudocode for the \DADEE{} algorithm is shown in Algorithm~\ref{alg:Train_DEE}.

\paragraph*{Uncertainty-aware CBF-based safe controller}
Having developed a new uncertainty estimation algorithm, we apply this algorithm to robot safety. Specifically, we use an uncertainty-aware CBF-based safe controller ~\cite{long2022safe,dhiman2023controlbarriers} as in \eqref{eq:CBF_opt}. We learn the unknown $F(\bfx)$ in the control affine system dynamics~\eqref{eq:system-dynamics}. We assume a fully observable state, which starts with a random initialization of system dynamics and high prior uncertainty.
We use $\epsilon$-greedy strategy~\cite{sutton2018reinforcement} for exploration exploitation trade-off.
When taking a random control $\bfu_r$ for $\epsilon$-greedy strategy, we find the closest-control $\|\bfu_k - \bfu_r\|_2$  that is safe according to \eqref{eq:socp-condition}.
As the robot explores the environment, we collect state $\bfx_t$ and control input $\bfu_t$ at each iteration in a replay buffer, which is used to learn the distribution of $F(\bfx)$ of unknown system dynamics.
The unknown $F(\bfx)$ are modeled by a 6 layer and 30 unit MLP with the same hyperparameters as in Section~\ref{sec:1-D-regression-experiment}.
We use various Bayesian learning algorithms to predict the expectation and variance of the $\mbox{CBC}$ as required by the controller in \eqref{eq:socp-condition}.
Then, we use \eqref{eq:CBF_opt} to find the optimal control input that is safe.
This algorithm is summarized in Algorithm~\ref{alg:DEE_CBF}.

\eqref{eq:socp-condition} can become infeasible when the variance grows faster than the expectation of $\CBC$ in all directions~\cite{castaneda2021pointwise}. This can occur when sufficient data is not available. 
When we encounter an infeasible optimization problem, we re-train the model on the same data to reduce the aleatoric uncertainty. If that does not work, we fall back to scaling uncertainty until the optimization becomes feasible.
\begin{algorithm}
\caption{\DADEE{} based CBF controller}\label{alg:DEE_CBF}
\begin{algorithmic}[1]
\State $\calB = []$ \Comment{Initialize replay buffer}
\While{$\|\bfx_t - \bfx_d\| < \delta$} \Comment{Do until reach goal $\bfx_d$}
\State $\calB.\text{append}(\bfx_t,\bfu_t)$
\State Learn $f(\bfx)$ and $G(\bfx)$ using \DADEE{} on buffer $\calB$
\State Find optimal safe control $\bfu_t$ using \eqref{eq:CBF_opt} with at most risk $p_k$ 
\State Apply $\bfu_t$ to the robot.
\EndWhile
\end{algorithmic}
\end{algorithm}%
\section{Experiments and results}
\label{sec:experiments}
\subsection{1-D regression experiment}
\label{sec:1-D-regression-experiment}
We evaluate the surveyed uncertainty estimation algorithms on a 1-D regression task to find the strengths and weaknesses of each approach. Our eventual aim is to select the best uncertainty estimation algorithm for safe robot controllers, but we begin by testing the algorithms in isolation, so that we have more control over both in-domain and out-of-domain sampling. Beginning with this 1-D regression task also minimizes the chances of other components of this complex pipeline influencing the results.

We sample a synthetic dataset from a function $f(x)=x^3/5-x+\epsilon$ where $\epsilon \in \calU[-0.5,0.5]$ is uniformly distributed. Test dataset contains 1000 sample points from $x_{\text{test}} \in \calX_{all} = [-3,3]$, and train dataset includes 513 sample data restricted to $x_{\text{train}} \in \calX_{in} = [-2.5, -0.75]\cup[0.75, 2.5]$.
The test data, and hence the evaluations, are split into two parts, in-domain $x_{\text{test}} \in \calX_{in}$ and out-of-domain (OOD) $x_{\text{test}} \in \calX_{all} \setminus \calX_{in}$. We deliberately oversample the noisy dataset in order to capture irreducible (aleatoric) uncertainty and to minimize the effect of epistemic uncertainty in in-domain data. 
In our experiments, in-domain uncertainty captures the aleatoric uncertainty and out-of-domain uncertainty captures the sum of both aleatoric and epistemic uncertainty.


We use the following hyperparameters for the chosen learning algorithms to learn the above function:
\begin{enumerate}
\item \emph{MC-Dropout}~\cite{gal2016dropout} is implemented with 20\% chance of dropping activations and 5 predictions are used to estimate the posterior distribution. \emph{MC-DropF} uses fixed drop units at inference time. This yield less sensitive predictions to the inputs.
\item \emph{SWAG}~\cite{maddox2019simple} stores the last 10 weights of the post-training trajectory  with learning rate 0.03 and draws 5 samples from the weight distribution to predict output.
\item \emph{Deep Ensembles}~\cite{lakshminarayanan2017simple} contains an ensemble of 5 models with randomly initialized weights and a shuffled dataset.
\item \emph{Anchored Ensembles}~\cite{pearce2020uncertainty} uses a prior of normal distribution with a regularizing factor $\lambda = 10.0$.
\item \emph{Laplace approximation}~\cite{daxberger2021laplace} uses the full matrix method to calculate the curvature of the loss. 
\item \emph{GP\_Sparse}~\cite{gpy2014} uses RBF kernel and 10\% of data as induced data points to approximate the matrix inversion.
\end{enumerate}

\begin{figure}
\centering
\includegraphics[width=0.50\linewidth,trim=0.1in 0.2in 0.0in 0.1in,clip]{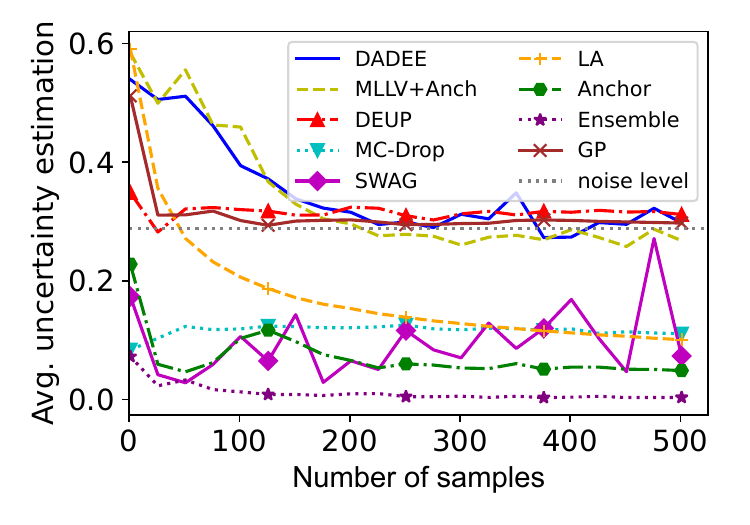}%
\includegraphics[width=0.50\linewidth,trim=0.1in 0.2in 0.0in 0.1in,clip]{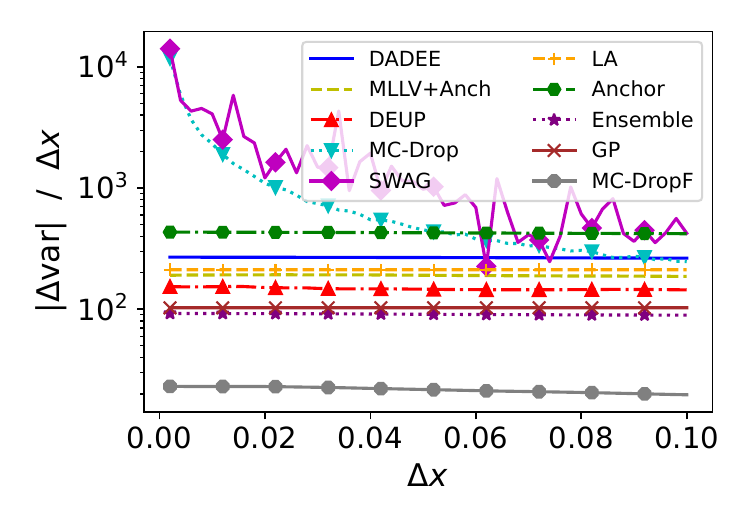}
\caption{ \textbf{(Left)} The variance estimated by Model-variance-based methods (MC-Dropout, SWAG, Anchor, Ensemble, and LA) approaches zero as the volume of training data increases. Meanwhile, the variance estimates produced by the direct estimation-based approaches (DEUP, MLLV) instead approach the true irreducible in-domain noise. \textbf{(Right)} Sensitivity in variance predictions: Sensitivity measures the change in variance as it responds to small changes in input (Defined in \eqref{eq:RelativeVariance}). MC-Dropout is unstable because a new set of activations is masked every time the network is reevaluated. }  
\label{fig:NoiseEstimation}
\label{fig:VarianceChange}
\end{figure}%
%
All the deep models in our study use an underlying neural network with 4 hidden layers, 10 hidden units and tanh activation functions. We deliberately choose an unusually high number of parameters for the toy example to replicate real-world neural networks where they are often overparameterized~\cite{zhang2021understanding}.

\subsubsection{Evaluation metrics}
To evaluate the accuracy of Bayesian learning algorithms in estimating uncertainty, we use
several standard metrics: test error (MSE), in-domain Mean Standardized Log Loss (MSLL)~\cite{williams2006gaussian}, out-of-domain MSLL, overall MSLL, in-domain root mean squared calibration error (RMSCE)~\cite{psaros2023uncertainty}, out-of-domain RMSCE, overall calibration error RMSCE, training time, inference time, memory allocated by method, and the number of trainable variables in the model. 
We do not use commonly used Brier score~\cite{brier1950verification} because it is specific to classification problems while we focus on a regression problem. We provide the formulation of MSLL~\cite{williams2006gaussian} and RMSCE~\cite{psaros2023uncertainty} here,
\begin{align*}
  \text{MSLL} \teq& \frac{1}{N}\sum_{i \in N}\frac{1}{2}\log(2\pi\sigma^2_{i,pred})+\frac{(\Bar{y}_{i,true}-\Bar{y}_{i,pred})^2}{2\sigma^2_{i,pred}}, \\
  \text{RMSCE} &\teq \sqrt{\frac{1}{|\calP|} \sum_{p_j \in \calP}\left[p_j - \frac{1}{N}\mathds{1}[ \bbP_{A}(\Bar{y}_{true}) \le p_j]\right]^2}, \label{eq:MSLL}
\end{align*}%
where $p_j \in \calP$ is chosen as a sequence of distribution intervals where we want to evaluate the uncertainty estimate, for example, $\calP = \{0, 0.01, 0.02, \dots, 1.0\}$. The second term in RMSCE, $\frac{1}{N}\mathds{1}[ \bbP_{A}(\Bar{y}_{true}) \le p_j]$ is the average fraction of observed points $\bar{y}_{true}$ that fall within the desired distribution interval $p_j$ according to the estimated uncertainty distribution $\bbP_{A}(.)$ by algorithm $A$. Here, $\mathds{1}$ is the indicator function and $N$ is the total number of sampled points. $\Bar{y}_{i,true}$ and $\sigma^2_{i,true}$ are mean and variance of $i$th sample point. MSLL can raise divide by zero error, when the $\sigma^2_{i,true}$ is close to zero. To avoid this, we added the $\epsilon = 0.01$ to variance predictions. We also report the accuracy of the learning algorithm itself using mean square error (\emph{MSELoss}).

The comparison of Bayesian learning algorithms is listed in Table~\ref{tab: Results of Bayesian Learning Algorithms}. We make several important observations.
GP\_Sparse~\cite{gpy2014,quinonero2005unifying} has a large train time due to overhead computations required to select inducing points, and is faster at inference time compared to GP. Although GP\_Sparse uses only 10 percent of data points, it produces the same validation loss and metric values. SWAG and MC-Dropout are fast in training and inference time but they are sensitive to input perturbations and require post hoc calibration. LA generates well-calibrated variance, but requires calculating large gradient matrices after each training, making it unsuitable for online learning. There are variations of LA that use low-rank matrices or last layer approximation, but they reduce the accuracy of prediction~\cite{daxberger2021laplace}. DEUP and Anchored Ensembles have complementary strengths, as well as being the most accurate approaches from their respective categories. We combine these two approaches into a new algorithm, which we call Direct Aleatoric with Deep Epistemic Ensembles (\DADEE). 

\DADEE{} has the lowest MSLL and RMSCE value for in-domain, out-of-domain, and overall test points among deep models. However, the training time of this model is longer than single deep models because it has more trainable parameters, and it has low inference time. 


The uncertainty estimates by four algorithms are visualized in Fig.~\ref{fig:1-D Example results}. The graphs represent the posterior distribution estimations generated by four selected approaches, namely GP, Anchored Ensembles, DEUP, and DADEE. These models were trained on the red data points and tasked with predicting the test dataset, including in-domain and out-of-domain (OOD) data. 
As stated previously, DEUP tends to underestimate the uncertainty associated with OOD data points. In contrast, model-variance-based methods can overestimate the OOD and could make the risk-aware controller conservative. 

We observe that model-variance-based algorithms do poorly in-domain and in an oversampled dataset. This is because the model-variance decreases arbitrarily with more data. This is expected due to central limit theorem. We demonstrate this observation in Figure~\ref{fig:NoiseEstimation}. The inclusion of direct estimators avoids this uncertainty collapse to zero on in-domain data.  

\subsubsection{Sensitivity to perturbations}
We also evaluate the uncertainty algorithms for sensitivity. We measure the sensitivity of an algorithm as the change in uncertainty estimates (variance) for small changes in inputs, 
\begin{equation}
  \label{eq:RelativeVariance}
  \text{Sensitivity} = \sum_{\bfx\in \calD} \frac{|\Var(\bfx + \Delta \bfx)- \Var(\bfx) |}{\Delta \bfx}.
\end{equation}%
The results are shown in Figure~\ref{fig:NoiseEstimation}. MC-Dropout and SWAG have the most sensitivity of all algorithms, while others have an acceptable level of sensitivity. We address the sensitivity issue of MC-Dropout by introducing MC-DropF where we freeze the activations to be dropped out at inference time.

\subsection{Robot Simulation}
\label{sec:robot-simulation}
\begin{figure}
\centering
\includegraphics[height=0.45\linewidth,trim=0.15in 0.15in 0.0in 0.2in,clip]{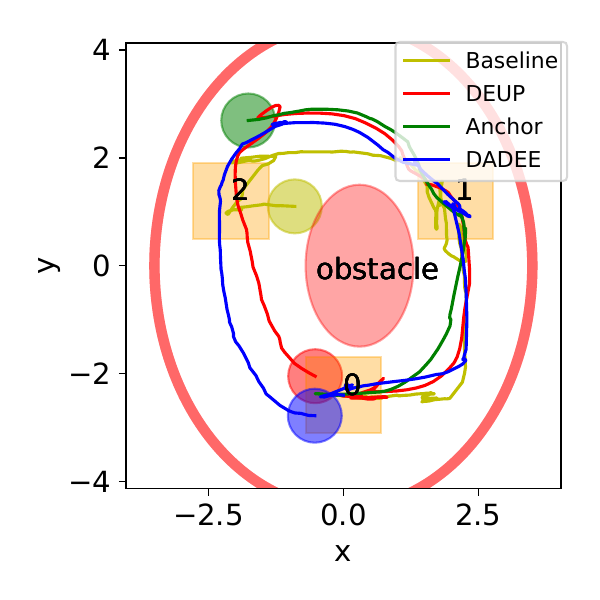}%
\raisebox{0.15in}{
\includegraphics[height=0.4\linewidth,trim=2.0in 0.0in 1.5in 0.5in,clip]{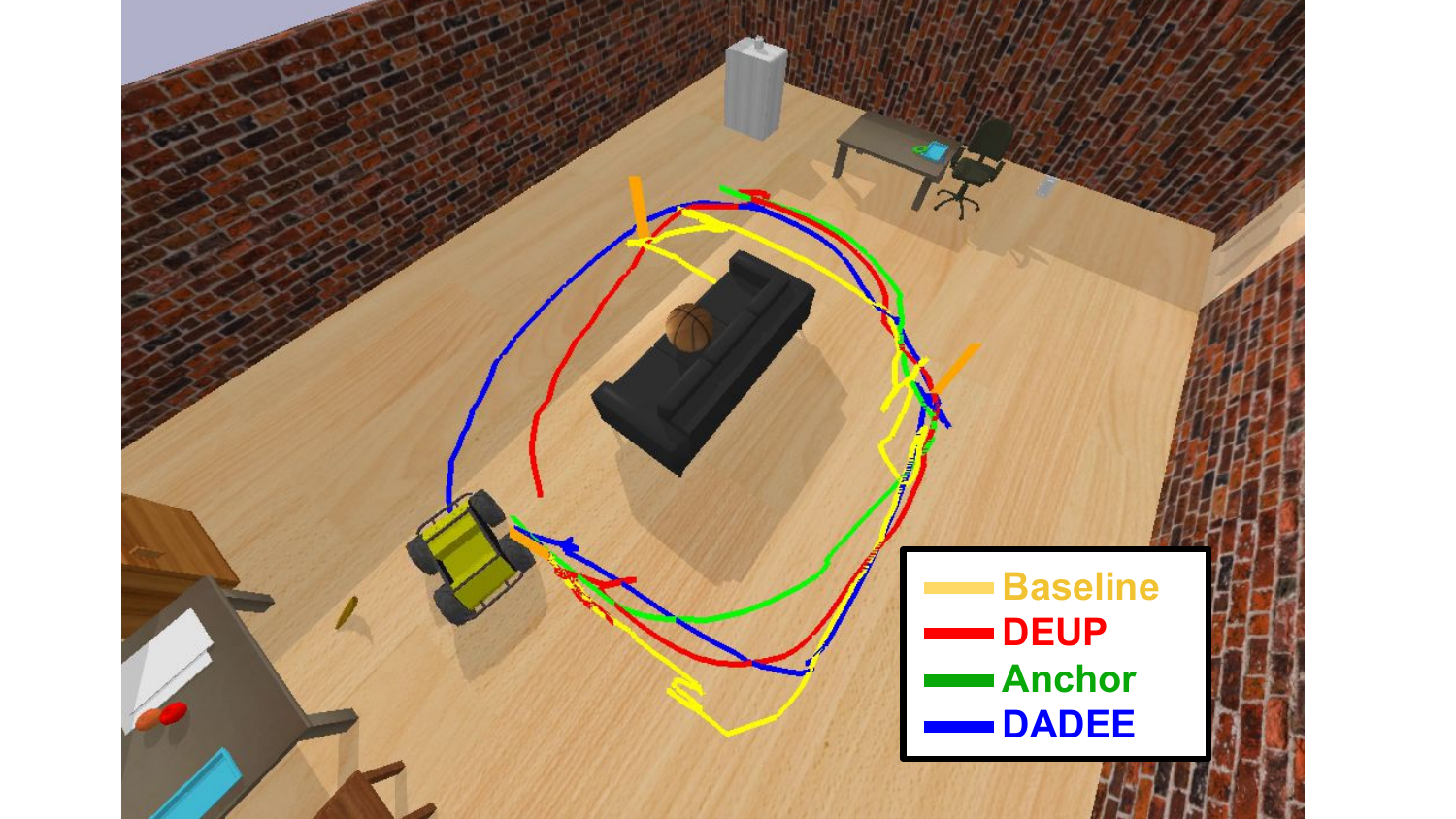}}
\caption{\textbf{(Left)} The trajectory of the simulated uncertainty-aware safe control-guided robot using Ensemble, DADEE, and MLP. \textbf{(Right)} 3D visualization of data shown left. }
\label{fig:PBModelTrajectory}
\label{fig:ModelTrajectory}
\end{figure}

In the previous section, we evaluated the uncertainty estimation of Bayesian learning algorithms, using the 1-D regression experiment. This allowed us to propose a new uncertainty estimation algorithm \DADEE{}. Now, we evaluate uncertainty estimation algorithms on safe-control task using Control Barrier Functions. We model obstacle avoidance as the safe-control task where the robot has to circle around an obstacle while avoiding an outer-wall obstacle as shown in Fig~\ref{fig:ModelTrajectory}\footnote{Video Link: \url{https://youtu.be/ttUpH-JBkHQ}. The video is generated using the control inputs that were computed offline.}. 
The simulations are conducted in a realistic PyBullet simulation environment with a Husky robot whose true complex system dynamics are unknown and are  learnt online.

The simulation environment, adapted from~\cite{bansal2020toolnet}, is a room with a couch in the middle. The task defined for the robot is to circle around the couch without hitting the room walls or the couch. We model the couch and the room as two elliptical control barrier conditions, $h_c(\bfx) \teq (\bfx-\bfx_c)^\top Q_c (\bfx-\bfx_c) - 1$ and $h_o(\bfx) \teq 1 - (\bfx - \bfx_o)^\top Q_o (\bfx - \bfx_o)$ respectively where $Q_c, \bfx_c, Q_o, \bfx_o$ are known parameters.
The elliptical trajectory around the couch is defined using three checkpoints shown in orange squares in Fig~\ref{fig:ModelTrajectory}. We use three checkpoints because two checkpoints leave ambiguity about the direction of rotation, while any more checkpoints would have avoided collision even case of straight lines.

We detail two parts of the safe control algorithm, (1) Bayesian learning and (2) safe control SOCP. We use the evaluated Bayesian learning algorithms to learn the system dynamics of the Husky robot.
To learn the system dynamics, we collect the states $\bfx$ (2D position and orientation) and applied controls $\bfu$ in a replay buffer, whose capacity is limited to 10000. 
Once the buffer is full the oldest observations are discarded.
Then, every 20 steps, we use the replay buffer to train the Bayesian learning model for 10 epochs.
Training can take a long time which can be unsafe for the robot.
This can be remedied by running the training loop in parallel on a different core of the CPU, whose implementation we have released in the Github repository. However, in these experiments, the training loop and control loop run sequentially. It took $0.410s \pm 0.047$ to run the control step and $0.941s \pm 0.099$ to run the training step on a i9-3.7 GHz 10-core CPU desktop.

At each time step, the following safe control SOCP optimization problem is solved:
{\small%
\begin{align}
  \min_{\bfu_k \in \bbR^m}  \quad &\| {\bfu_k}\|_2^2  + \lambda \nabla_{\bfx_k}^\top V(\bfx_k, \bfx_d) F(\bfx_k) \ubfu_k \notag \\
  \text{s.t. } \quad & 
  \bbP(\CBC_c \ge \zeta|\bfx_k, \bfu_k) \ge \tilde{p}_k, \quad
  \bbP(\CBC_r \ge \zeta|\bfx_k, \bfu_k) \ge \tilde{p}_k, \notag \\
  &\bfu_{min} \le \bfu_k \le \bfu_{max}
  \label{eq:robot_exp_opt}.
\end{align}%
}
The control input $\bfu \teq [ v, \omega]$ contains the linear $v$ and angular velocity $\omega$, and the function $V$ is defined as $V(\bfx_k, \bfx_d) \teq (\bfx_k - \bfx_d)^2$ where $\bfx_d$ is the state of one of the three target checkpoint, $d=\{1, 2, 3\}$. $\CBC_c$ and $\CBC_o$ are the control barrier conditions for the couch and the room obtained from $h_c$ and $h_o$ using \eqref{eq:CBCdef}, respectively. Once the robot approaches the desired checkpoint by a given threshold, it is assigned to next one in a circular order $d = 1\to2\to3\to1$. 
We find that CBFs, $h_c$ and $h_r$, have a relative degree two in angular velocity $\omega$. To address this, we use the Constraint Transformation~\cite[Sec~III.C]{xiao2022control} to convert the CBFs to relative degree 1. 

We evaluate a selection of four Bayesian learning algorithms for comparison of safe control algorithms,
(1) \texttt{Baseline}, where we ignore uncertainty by setting $c(\tilde{p}_k) = 0$ in \eqref{eq:socp-condition}, (2) Anchored Ensembles~\cite{pearce2020uncertainty}, (3) DEUP~\cite{lahlou2021deup}, and (4) \DADEE{}. The trajectories for robot movement using these methods are visualized in Fig~\ref{fig:ModelTrajectory}. 
For each trajectory, the unknown system dynamics is being learned online and $p_k$ is set to 0.95. The jagged trajectories are due to randomized safe controls using an $\epsilon$-greedy algorithm. We recall from Sec.~\ref{sec:1-D-regression-experiment} that Anchored Ensemble only predicts the OOD uncertainty, and DEUP captures aleatoric uncertainty, and \DADEE{} accounts for both in-domain and OOD uncertainty. \DADEE{} is more conservative and stays away from the obstacle when its uncertainty is high.

The error rate of all four methods along with MC-Dropout~\cite{gal2016dropout} are shown in Table~\ref{tab:error-rate}.
To compute the error rate, we run the Husky robot 20 times around the couch
for different acceptance risks $p_k$. We define the error rate as the number of failed steps divided by the total number of steps when any of the control barrier conditions were active in the optimization of \eqref{eq:robot_exp_opt}. The error rate results are summarized in Table~\ref{tab:error-rate}. We find that although uncertainty estimation algorithms were well calibrated as seen in Sec.~\ref{sec:1-D-regression-experiment}, the error rate is much more conservative than acceptable risk $p_k$. For example, at $p_k = 0.9$ the error rate of MC-Dropout is $0.03$ which is conservative by a factor 3. The failure rate of \DADEE{} is lower than in other methods, demonstrating its safety both in-domain and out-of-domain.

\begin{algorithm*}
\caption{Train \DADEE}\label{alg:Train_DEE}
\begin{algorithmic}[1]
\State Input: a dataset $\calD = \{(\bfx_1, \bfy_1), \dots, (\bfx_n, \bfy_n)\}$,  
Neural network architecture $f(\bfx; \theta)$ with unknown weights
\State Input: Prior distribution $\pi(\theta)$ over the weights
\State $\text{Initialize ensemble weights } \beta_1, \dots \beta_L \text{, and direct estimator weights } \beta_v$ and variance dataset $\calE = \{\}$
\State Sample anchor weights $\theta_a \sim \pi(\theta)$ from the prior
\For{$l\gets 1\dots L$} \Comment{Repeat for each network in the ensemble}
    \State Train Anchored ensemble networks to find $\beta^*_l$ using \eqref{eq:MSELoss} on dataset $\calD$
\EndFor
\For{$l\gets 1\dots L$}
    \State $e_{i} = \|\frac{1}{L} \sum_{l=1}^L f(\bfx_i; \beta_l^*)-\bfy_i\|^2 $ for all $(\bfx_i, \bfy_i) \in \calD$ \Comment{Compute error per sample}
    \State $\calE = \calE \cup \{(\bfx_i, e_i)\}$ for all $(\bfx_i) \in \calD$ \Comment{Add errors to the variance dataset}
\EndFor
\State $\beta^*_v = \arg~\min_{\beta_v} \sum_{i \in \calE} \|f(\bfx_i; \beta_v) - e_i\|_2^2$  \Comment{Train a direct estimator of variance }
\State Return mean $\mu(\bfx) = \frac{1}{L} \sum_{l=1}^L f(\bfx;\beta_l^*)$
and variance $\sigma^2(\bfx) = f(\bfx; \beta_v) + \frac{1}{L-1} \sum_{l=1}^L \|\mu(\bfx) - f(\bfx;\beta_l^*)\|_2^2$
\end{algorithmic}
\end{algorithm*}

\begin{table}
    \centering
    \caption{Error rate of different methods in combination with CBF in a realistic simulation. \DADEE{}  leads to a safer result than anchored ensembles. The error rate when no uncertainty is considered is 0.242.}
    \label{tab:error-rate}
\begin{tabular}{@{}lccccc@{}}
\toprule
\multirow{2}{*}{Model} & \multicolumn{5}{c}{$p_k$}      \\ \cmidrule(l){2-6} 
                       & 0.5 & 0.6 & 0.7 & 0.8 & 0.9 \\ \midrule
        MC-Dropout~\cite{gal2016dropout}   & 0.089  & 0.039 & 0.042 & 0.037 & 0.037 \\
        Anchored~\cite{pearce2020uncertainty}   & 0.058  & 0.033 & 0.023 & 0.010 & 0.004 \\
        DEUP~\cite{lahlou2021deup}   & 0.217  & 0.087 & 0.033 & 0.012 & 0.009 \\
        DADEE & 0.044 & 0.043 & 0.027 & 0.008 & 0.003 \\
        \bottomrule
\end{tabular}
\end{table}

        



\section{Conclusion and Future work}
Our experiments show that the existing uncertainty estimation algorithms are not able to accurately estimate both in-domain and out-of-domain uncertainty, and that a combination of in-domain uncertainty and out-of-domain uncertainty is needed in safety critical applications.
In future, we plan to further develop this work by applying it beyond simulation, to real robots.

\bibliographystyle{IEEEtran}
\bibliography{IEEEabrv,bib/main}
\end{document}